\documentclass{article}
\usepackage{subcaption}
\usepackage{mathtools}
\usepackage{microtype}
\usepackage{bbm}
\usepackage{makecell}
\usepackage[capitalise]{cleveref}

\usepackage{amsfonts}
\usepackage{amsmath}
\usepackage{IEEEtrantools}
\usepackage{amssymb} 

\usepackage[noend]{algpseudocode}

\usepackage{xcolor}
\usepackage{color}
\usepackage{graphicx}

\usepackage{amsmath}

\usepackage{float}

\usepackage[normalem]{ulem}
\usepackage{amsmath, amssymb, epsfig, bm}
\usepackage{graphicx}
\usepackage{caption}
\usepackage{mathrsfs}
\usepackage[linesnumbered,ruled,lined]{algorithm2e}
\SetKwInput{KwInput}{Input}
\SetKwInput{KwOutput}{Output}
\newcommand{\myfootnote}[1]{
\renewcommand{\thefootnote}{}
\footnotetext{\hspace{-16.5pt}\footnotesize#1}
\renewcommand{\thefootnote}{\arabic{footnote}}}

\def \P {\mathbb{P}}

\def \x {\vct{x}}

\def\vx{{\bm{x}}}

\def\mA{{\bm{A}}}
\def\mB{{\bm{B}}}
\def\mC{{\bm{C}}}

\def\mH{{\bm{H}}}

\def\mL{{\bm{L}}}

\def\mW{{\bm{W}}}
\def\mX{{\bm{X}}}
\def\mY{{\bm{Y}}}
\def\mZ{{\bm{Z}}}
\def\m\mY{{\bm{\mY}}}
\def\m\mZ{{\bm{\mZ}}}

\newcommand{\tran}{^{\top}}
\DeclareMathOperator{\tr}{tr}
\DeclareMathOperator{\avg}{{avg}}
\usepackage[top=0.5in, bottom=0.75in, left=1in, right=1in]{geometry}
\usepackage{amsmath}

\begin{document}
\title{Guided Semi-Supervised Non-negative Matrix Factorization on Legal Documents}

\author{Pengyu Li  $^\dagger$ \textsuperscript{\rm 1}, 
        Christine Tseng $^\dagger$ \textsuperscript{\rm 1}, 
        Yaxuan Zheng $^\dagger$ \textsuperscript{\rm 1}, \\
        Joyce A. Chew \textsuperscript{\rm 1}, 
        Longxiu Huang \textsuperscript{\rm 1}, 
        Benjamin Jarman \textsuperscript{\rm 1}, 
        Deanna Needell \textsuperscript{\rm 1}
}

\myfootnote{\noindent \textsuperscript{\rm 1} Department of Mathematics, University of California, Los Angeles, CA \\
    $\dagger$: These authors contributed equally to this work.\\
    Email addresses: erby1215@g.ucla.edu (P.L.), christinetseng@g.ucla.edu (C.T.), yaxuan@g.ucla.edu (Y.Z.), joycechew@math.ucla.edu (J.A.C.), huangl3@math.ucla.edu (L.H.), bjarman@math.ucla.edu (B.J.),  deanna@math.ucla.edu (D.N.) 
}

\maketitle

\begin{abstract}
    
Classification and topic modeling are
popular techniques in machine learning that extract information from large-scale datasets. By incorporating \textit{a priori} information such as labels or important features, methods have been developed to perform classification and topic modeling tasks; however, most methods that can perform both do not allow for guidance of the topics or features. In this paper, we propose a method, namely Guided Semi-Supervised Non-negative Matrix Factorization (GSSNMF), that performs both classification and topic modeling by incorporating supervision from both pre-assigned document class labels and user-designed seed words. We test the performance of this method through its application to legal documents provided by the California Innocence Project, a nonprofit that works to free innocent convicted persons and reform the justice system. The results show that our proposed method improves both classification accuracy and topic coherence in comparison to past methods like Semi-Supervised Non-negative Matrix Factorization (SSNMF) and Guided Non-negative Matrix Factorization (Guided NMF).
\end{abstract}

\section{Introduction}

Understanding latent trends within large-scale, complex datasets is a key component of modern data science pipelines, leading to downstream tasks such as classification and clustering. In the setting of textual data contained within a collection of documents, non-negative matrix factorization (NMF) has proven itself as an effective, unsupervised tool for this exact task \cite{lee1999learning, lee2001algorithms, arora12learningtopic, kuang2015nonnegative, xu2003document}, with trends represented by topics. 
Whilst such fully unsupervised techniques bring great flexibility in application, it has been demonstrated that learned topics may not have the desired effectiveness in downstream tasks \cite{chang2009reading}. In particular, certain related features may be strongly weighted in the data, yielding highly related topics that do not capture the variety of trends effectively \cite{jagarlamudi2012incorporating}.
To counteract this effect, some level of supervision may be introduced to steer learnt topics towards being more meaningful and representative, and thus improve the quality of downstream analyses. Semi-Supervised NMF (SSNMF) \cite{lee2009semi, chen2008non, haddock2020semisupervised} utilizes class label information to simultaneously learn a dimensionality-reduction model and a model for a downstream task such as classification. Guided NMF \cite{vhrn20gNMF} instead incorporates user-designed seed words to ``guide" topics towards capturing a more diverse range of features, leveraging (potentially little) supervision information to drive the learning of more balanced, distinct topics. 
Despite a distinction between the supervision information and goals of Guided NMF and SSNMF, there are certainly mutual relationships: knowledge of class labels can be leveraged to improve the quality of learnt topics (in terms of scope, mutual exclusiveness, and self-coherence), whilst \textit{a priori} seed words for each class can improve classification.

With these heuristics in mind, we introduce Guided Semi-Supervised NMF (GSSNMF), a model that incorporates both seed word information and class labels in order to simultaneously learn topics and perform classification. The goal of this work is to show that utilizing both forms of supervision information concurrently offers improvements to both the topic modeling and classification tasks, while producing highly interpretable results.

\section{Related Work}
\subsection{Classical Non-negative Matrix Factorization}
Non-negative matrix factorization (NMF) is a powerful framework for performing unsupervised tasks such as topic modeling and  clustering \cite{lee1999learning}.
Given a target dimension $k<\min\{d,n\}$, the classical NMF method approximates  a non-negative data matrix $\mX\in\mathbb{R}^{d\times n}$ by the product of  two   non-negative low-rank  matrices: the \emph{dictionary matrix} $\bm{W} = [\mathbf{w}_1, \mathbf{w}_2, \cdots, \mathbf{w}_k] \in \mathbb{R}_{\geq 0}^{d \times k}$ and the \emph{coding matrix} $\bm{H} = [\mathbf{h}_1, \mathbf{h}_2, \cdots, \mathbf{h}_n] \in \mathbb{R}_{\geq 0}^{k \times n}$.
Both matrices $\bm{W}$ and $\bm{H}$ can be found by solving the optimization problem 
\begin{equation}\label{eq: c_NMF_eq}
\operatorname*{argmin}_{\mW\in \mathbb{R}^{d\times k}_{\ge 0},\, \mH\in \mathbb{R}^{k\times n}_{\ge 0} } \frac{1}{2} \lVert \mX - \mW\mH  \rVert_{F}^{2},
\end{equation}
where $\|\bm{A}\|_F^2 = \sum_{i,j}a_{ij}^2$ is  the Frobenius norm of $\bm{A}$ and the constant of $\frac{1}{2}$ is to ease the calculation when taking the gradient. In the context of topic modeling, the dimension $k$ is the number of desired topics. Each column of $\bm{W}$ encodes the strength of every dictionary word's association with a learned topic, and each column of $\bm{H}$ encodes the relevance of every topic for a given document in the corpus. By enforcing non-negativity constraints, NMF methods can learn topics and document classes with high interpretability \cite{lee1999learning, xu2003document}. 

\subsection{Semi-Supervised NMF} \label{subsec:subsec_ssnmf}
Beside topic modeling, one variant of classical NMF, Semi-Supervised NMF (SSNMF) \cite{lee2009semi, chen2008non, haddock2020semisupervised}, is designed to further perform classification. SSNMF introduces a masking matrix $\mL=[\mathbf{\ell}_1,\dots,\mathbf{\ell}_n]\in\mathbb{R}_{\geq 0}^{p\times n}$ and a label matrix $\mZ=[\mathbf{z}_1, \cdots, \mathbf{z}_n]\in \mathbb{R}_{\geq 0}^{p\times n}$, where $p$ is the number of classes and $n$ is the number of documents. The masking matrix $\mL$ is defined as:
\begin{equation}
    \mathbf{\ell}_j = \begin{cases}
    \textbf{1}_p, & \text{if the label of document $x_j$ is known}\\
    \textbf{0}_p, & \text{otherwise},
    \end{cases} 
\end{equation}
where $\textbf{1}_p = [1,\dots,1]^T\in \mathbb{R}^p$ and $\textbf{0}_p = [0,\dots,0]^T\in \mathbb{R}^p$. Note that the masking matrix $\mL$ splits the label information into train and test sets. Each column $\mathbf{z}_i$ is a binary encoding vector such that if the document $\vx_i$ belongs to  class $j$, the $j^{th}$ entry of $\mathbf{z}_i$ is $1$ and otherwise it is set to be $0$. The dictionary matrix $\mW$, coding matrix $\mH$, and label dictionary matrix $\mC$ can be found by solving the optimization problem
\begin{equation}\label{eq:ss_NMF_eq}
\operatorname*{argmin}_{\mW \in \mathbb{R}^{d \times k}_{\geq 0}, \mH \in \mathbb{R}^{k\times n}_{\geq 0},\mC \in \mathbb{R}^{ p\times k }_{\geq 0}} \underbrace{\frac{1}{2}\|\mX -\mW \mH\|^2_F}_{\text{classical NMF}} + \underbrace{\frac{\mu}{2} \|\mL \odot (\mZ - \mC \mH)\|^2_F}_{\text{label information}}, 
\end{equation}
where $\mu > 0$ is a regularization parameter and $\mA \odot \mB$ denotes entry-wise multiplication between matrix $\mA$ and $\mB$. Matrices $\mW$ and $\mH$ can be interpreted in the same way as classical NMF. Matrix $\mC$ can be viewed as the dictionary matrix for the label matrix $\mZ$.

\subsection{Guided NMF} \label{subsec:subsec_GuidedNMF}
Due to the fully unsupervised nature of classical NMF, 
the {generated} topics 
may suffer from redundancy or lack of cohesion when the given data set is biased towards a set of featured words \cite{chang2009reading, jagarlamudi2012incorporating, vhrn20gNMF}. Guided NMF \cite{vhrn20gNMF} addresses this by guiding the topic outputs through incorporating flexible user-specified \textit{seed word} supervision. Each word in a given list of $s$ seed words can be represented as a sparse binary \textit{seed vector} $\mathbf{v} \in \mathbb{R}^{d}$, whose entries are zero except for some positive weights at entries corresponding to the seed word feature. The corresponding binary \textit{seed matrix} $\mY \in \mathbb{R}^{d \times s}_{\geq0}$ can be constructed as
\begin{align*}
    \mY = [\mathbf{v_1}, \cdots, \mathbf{v_s}] \in \mathbb{R}^{d \times s}_{\geq0}.
\end{align*}
For a given data matrix $\mX$ and a seed matrix $\mY$, Guided NMF seeks a dictionary matrix $\mW$, coding matrix $\mH$, and topic supervision matrix $\mB$ by considering the optimization problem
\begin{equation}\label{eq:g_NMF_eq}
\operatorname*{argmin}_{\mW \in \mathbb{R}^{d \times k}_{\geq 0},\mH \in \mathbb{R}^{k\times n }_{\geq 0},\mB \in \mathbb{R}^{k \times s}_{\geq 0}} \underbrace{\frac{1}{2}\|\mX - \mW\mH\|^2_F}_{\text{classical NMF}} + \underbrace{\frac{\lambda}{2}\|\mY - \mW\mB\|^2_F}_{\text{guiding}},
\end{equation}
where $\lambda > 0$ is a regularization parameter. Matrices $\mW$ and $\mH$ can be interpreted in the same way as classical NMF. Matrix $\mB$ can help identify topics that form from the influence of seed words.

\section{Proposed Methods} \label{sec:GSSNMF}

Recall that SSNMF is able to classify different documents through given label information, while Guided NMF can guide the content of generated topics via \textit{a priori} seed words. We propose a {more general model}, 
Guided Semi-Supervised NMF (GSSNMF), that can leverage both label information and important seed words to improve performance in both multi-label classification and topic modeling. Heuristically, we see that for classification, the user-specified seed words aid SSNMF in distinguishing between each class label and thus improve classification accuracy. For topic modeling, the 
known label information enables Guided NMF to better cluster similar documents, improving topic coherence and interpretability. In particular, GSSNMF optimizes
\begin{equation}\label{eq:GSSNMF_loss_function}
    \operatorname*{argmin}_{\mW \in \mathbb{R}^{d\times k}_{\geq 0}, \mH \in \mathbb{R}^{k\times n}_{\geq 0}, \mB \in \mathbb{R}^{k\times s}_{\geq 0}, \mC \in \mathbb{R}^{p \times k}_{\geq 0}} \underbrace{\frac{1}{2}\|\mX - \mW \mH\|_F^2}_{\text{classical NMF}} + \underbrace{\frac{\lambda}{2}\|\mY -\mW \mB\|_F^2}_{\text{guiding}} + \underbrace{\frac{\mu}{2}\|\mL \odot (\mZ - \mC \mH)\|_F^2}_{\text{label information}},
\end{equation} 
where matrices $\mW$, $\mH$, $\mY$, $\mB$, $\mL$, $\mZ$, and $\mC$ and constants $\lambda$ and $\mu$ can be interpreted in the same way as in SSNMF and Guided NMF. Note that if we simply set either $\lambda$ or $\mu$ to be $0$, GSSNMF reduces to SSNMF or Guided NMF respectively. 

To solve \eqref{eq:GSSNMF_loss_function}, we propose a multiplicative update scheme akin to those in \cite{4359171, lee2009semi}. The derivation of these updates is provided in \cref{app:GSSNMF_algorithm_proof}, and the updating process is presented 
in Algorithm \ref{alg:GSSNMF_algorithm}.

\begin{center}
\begin{minipage}{0.77\linewidth}
\IncMargin{0.5em}
\begin{algorithm}[H]
\caption{GSSNMF with multiplicative updates}
\label{alg:GSSNMF_algorithm}
\DontPrintSemicolon
\SetAlgoLined
\KwInput{$\mX \in \mathbb{R}^{d\times n}_{\geq0}$, $\mY \in \mathbb{R}^{d\times s}_{\geq0}$, $\mL \in \mathbb{R}^{p\times n}_{\geq 0}$, $\mZ \in \mathbb{R}^{p\times n}_{\geq0}$, $k$, $\lambda$, $\mu$, $N$}
Initialize $\mW \in \mathbb{R}^{d\times k}_{\geq0}, \mH \in \mathbb{R}^{k\times n}_{\geq0}, \mB \in \mathbb{R}^{k\times s}_{\geq 0}, \mC \in \mathbb{R}^{p\times k}_{\geq0}$\;  
\For{$i = 1, \ldots, N$}    
    	{$\mW \leftarrow \mW\odot \frac{\mX\mH\tran +\lambda \mY\mB\tran}{\mW\mH\mH\tran + \lambda \mW\mB\mB\tran}$ \\
    	$\mH \leftarrow \mH\odot \frac{\mW\tran \mX + \mu C\tran (\mL\odot \mL \odot \mZ)}{\mW\tran \mW\mH + \mu \mC\tran(\mL\odot \mL\odot \mC\mH)}$ \\
    	$\mB \leftarrow \mB\odot \frac{\mW\tran \mY}{\mW\tran \mW\mB}$ \\
    	$\mC \leftarrow \mC\odot \frac{(\mL\odot \mL \odot \mZ)\mH\tran}{(\mL \odot \mL \odot \mC\mH)\mH\tran}$}
\KwOutput{$\mW, \mH, \mB, \mC$}
\end{algorithm}
\DecMargin{1em}
\end{minipage}
\end{center}

We will demonstrate the strength of GSSNMF by considering different combinations of  the parameters $\lambda$ and $\mu$, and comparing the implementations of this method with SSNMF and Guided NMF through real-life applications in the following section. 

\section{Experiments} \label{sec:exp}

In this section, we evaluate the performance of our GSSNMF on the California Innocence Project dataset \cite{budahazy2021analysis}. Specifically, we compare GSSNMF with SSNMF for performance in classification (measured by the Macro-F1 score, i.e the averaged F1-score which is sensitive to different distributions of different classes \cite{opitz2019macro}) and  with Guided NMF for performance in topic modeling (measured by the $\mathcal{C}$ coherence score \cite{mimno-etal-2011-optimizing}). 

\subsection{Data and Pre-Processing} \label{subsec:data}

 A nonprofit, clinical law school program hosted by the California Western School of Law, the California Innocence Project (CIP) focuses on freeing wrongfully-convicted prisoners, reforming the criminal justice system, and training upcoming law students \cite{budahazy2021analysis}. Every year, the CIP receives over 2000+ requests for help, each containing a case file of legal documents. Within each case file, the \textit{Appellant’s Opening Brief (AOB)} is a legal document written by an appellant to argue for innocence by explaining the mistakes made by the court. This document contains crucial information about the crime types relevant to the case, as well as potential evidence within the case \cite{budahazy2021analysis}. For our final dataset, we include all AOBs in case files that have assigned crime labels, totaling $203$ AOBs. Each AOB is thus associated with one or more of thirteen crime labels: \textit{assault}, \textit{drug}, \textit{gang}, \textit{gun}, \textit{kidnapping}, \textit{murder}, \textit{robbery}, \textit{sexual}, \textit{vandalism}, \textit{manslaughter}, \textit{theft}, \textit{burglary}, and \textit{stalking}.
 
To pre-process data, we remove numbers, symbols, and stopwords according to the NLTK English stopwords list \cite{bird2009natural} from all AOBs; we also perform stemming to reduce inflected words to their original word stem. Following the work of \cite{ramos2003using, li2007keyword, budahazy2021analysis}, we apply \textit{term-frequency inverse document frequency (tf-idf)}  \cite{salton1988term} to our dataset of AOBs and generated the corpus matrix $\mX$ with parameters \texttt{max\_df = 0.8}, \texttt{min\_df=0.04}, and \texttt{max\_features = 700} in the function \texttt{TfidfVectorizer}. 
 
\begin{figure}[th]
    \centering
    \includegraphics[scale=0.65]{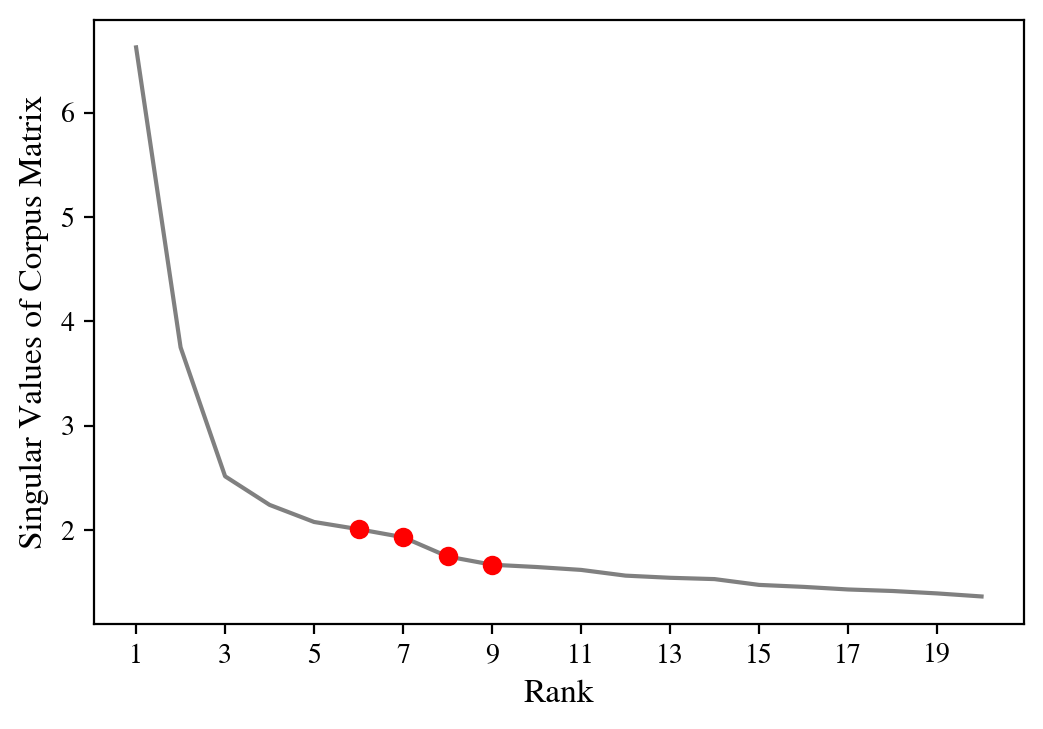}
    \caption{First $20$ singular values of the corpus matrix $\mX$.}
    \label{fig:TM_elbow_plot}
\end{figure}

For our topic modeling methods, we also need to identify the number of topics that potentially exist in our data, which corresponds to the rank of corpus matrix $\mX$. To determine the proper range of the number of topics to generate, we analyze the singular values of $\mX$. The magnitudes of singular values are positively related to the increment in proportion variance explained by adding on more topics, or increasing rank, to split the corpus matrix $\mX$ \cite{grotheer2020covid19}. In this way, we use the number of singular values to approximate the rank of $\mX$. Figure \ref{fig:TM_elbow_plot} plots the magnitudes of the singular values of corpus matrix $\mX$ against the number of singular values, which is also the approximated rank. By examining this plot, we see that a range for potential rank is between $6$ to $9$, since the magnitudes of the singular values start to level off around this range.  
 
\subsection{GSSNMF for Classification}

Taking advantage of Cross-Validation methods \cite{stone1974cross}, we randomly split all AOBs into a $70 \%$ training set with  labeled information and a $30 \%$ testing set without labels. In practice, $70 \%$ of the columns of the masking matrix $\mL$ are set to $\textbf{1}_p$ for the training set, and the rest are set to $\textbf{0}_p$ for the testing set. As a result, the label matrix $\mZ$ is masked by $\mL$ into a corresponding training label matrix $\mZ_{\text{train}}$ and a corresponding testing label matrix $\mZ_{\text{test}}$. We then perform SSNMF and GSSNMF to reconstruct $\mZ_{\text{test}}$, setting the number of topics, or rank, equal to $8$. Given the multi-label characteristics of the AOBs, we compare the performance between SSNMF and GSSNMF with a measure of classification accuracy: the \textit{Macro F1-score}, which is designed to access the quality of multi-label classification on $\mZ_{\text{test}} \cite{opitz2019macro}$. The Macro F1-score is defined as 
\begin{equation}
    \mathrm{Macro \ F1 \mbox{-} score} = \frac{1}{p} \sum_{i = 0}^p \mathrm{F1\mbox{-} score}_i,
\end{equation}
where $p$ is the number of labels and F1-score$_i$ is the F1-score for topic $i$. Notice that the Macro F1-score treats each class with equal importance; thus, it will penalize models which only perform well on major labels but not minor labels. In order to handle the multi-label characteristics of the AOB dataset, we first extract the number of labels assigned to each AOB in the testing dataset. Then for each corresponding column $i$ of the reconstructed $\mZ_{\text{test}}$, we set the largest $j_i$ elements in each column to be $1$ and the rest to be $0$, where $j_i$ is the true number of labels assigned to the $i$th document in the testing set. 

\begin{figure}[H]
    \centering

  \begin{subfigure}[t]{0.15\textwidth}
        \raisebox{.087\height}{\includegraphics[scale=0.1395]{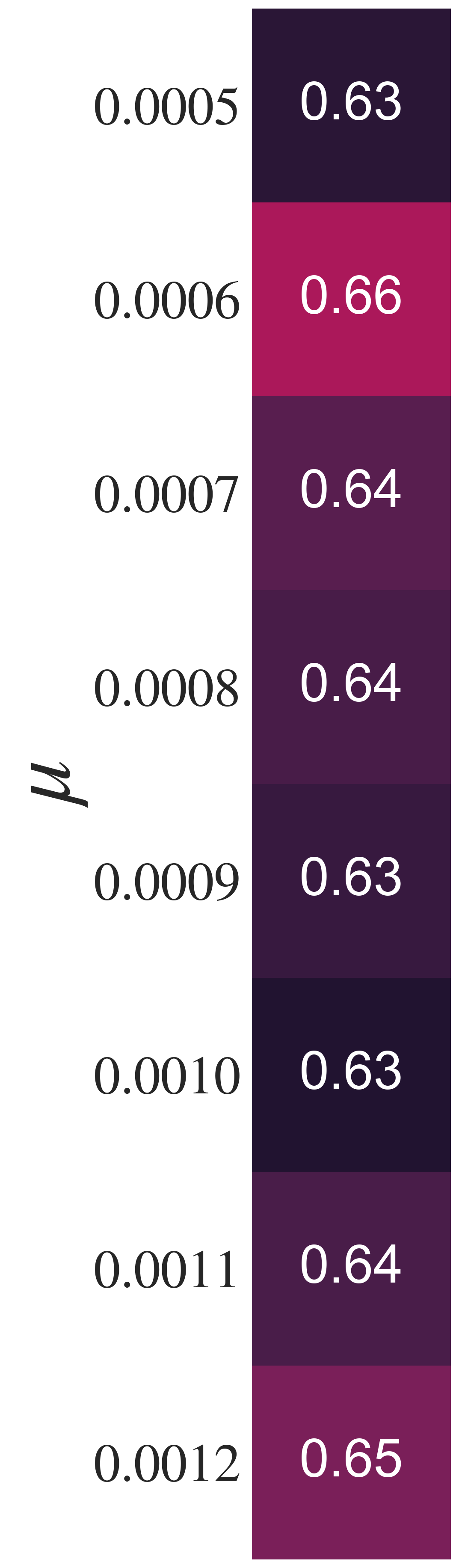}}
        \caption{SSNMF}
        \label{fig:SSNMG_col}
  \end{subfigure}
  ~
  \begin{subfigure}[t]{0.17\textwidth}
        \raisebox{.087\height}{\hspace{20pt}\includegraphics[scale=0.1395]{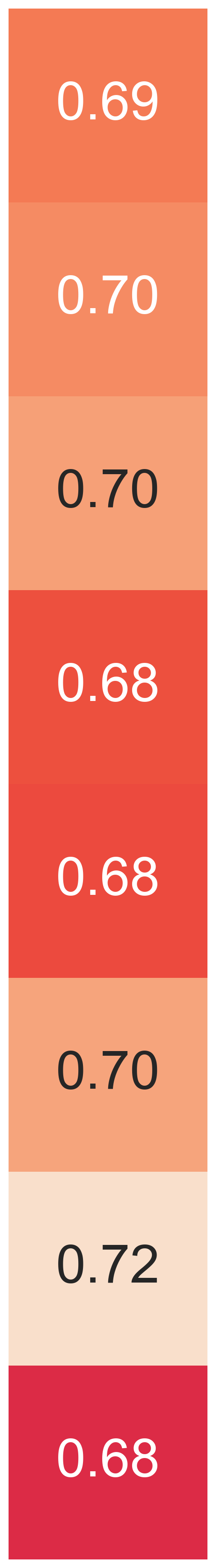}}
        \caption{GSSNMF}
        \label{fig:GSSNMF_hm_col}
  \end{subfigure}
 ~
  \begin{subfigure}[t]{0.3\textwidth}
        \includegraphics[scale=0.21]{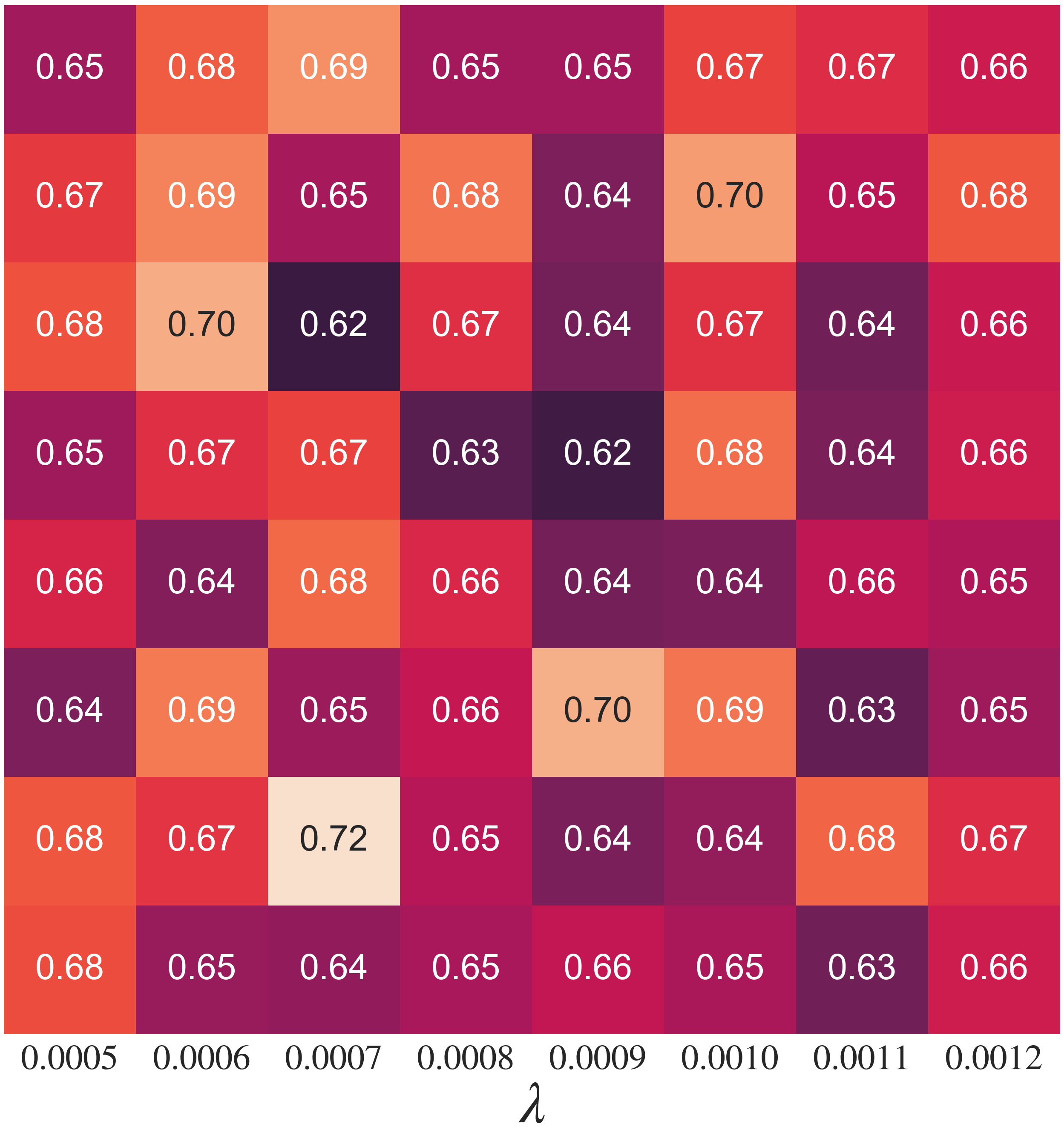}
        \caption{GSSNMF}
        \label{fig:GSSNMF_hm}
  \end{subfigure}
  ~
  \centering
    \begin{subfigure}[t]{0.5\textwidth}
        \includegraphics[scale=0.5]{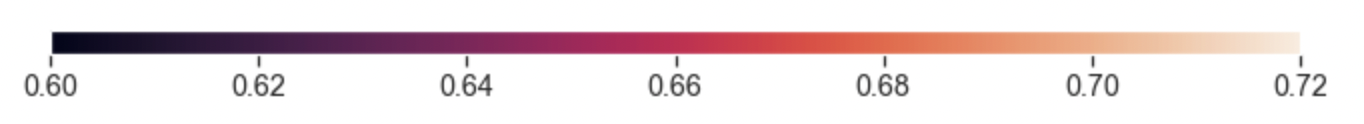}
  \end{subfigure}
  
  \caption{The heatmap representation of Macro F1-score averaged over 10 independent trials for a proper range of $\mu$. (\textbf{a}) is the set of highest SSNMF Macro F1-score. (\textbf{b}) is how GSSNMF improves on SSNMF results in (\textbf{a}) using a proper $\lambda$. (\textbf{c}) is GSSNMF Macro F1-score for a proper range of $\lambda$ tested with each $\mu$. Note that (\textbf{b}) is the row maximums of (\textbf{c}).}
  \label{fig:heat_map_all}
\end{figure}

We first tune the parameter of $\mu$ in SSNMF to identify a proper range of $\mu$ for which SSNMF performs the best on the AOB dataset under the Macro F1-score. Then for each selected $\mu$ in the proper range, we run GSSNMF with another range of $\lambda$. While there are various choices of seed words, we naturally pick the class labels themselves as seed words for our implementation of GSSNMF. 
As a result, for each combination of $\mu \in [0.0005, 0.0006, \cdots, 0.0012]$ and 
$\mu, \lambda \in [0.0005, 0.0006, \cdots, 0.0012]$, we conduct $10$ independent Cross-Validation trials and  average the Macro F1-scores. The results are displayed 
in Figure \ref{fig:heat_map_all}.

\begin{figure}[H]
    \centering
  \begin{subfigure}[t]{0.8\textwidth}
        \includegraphics[scale=0.4]{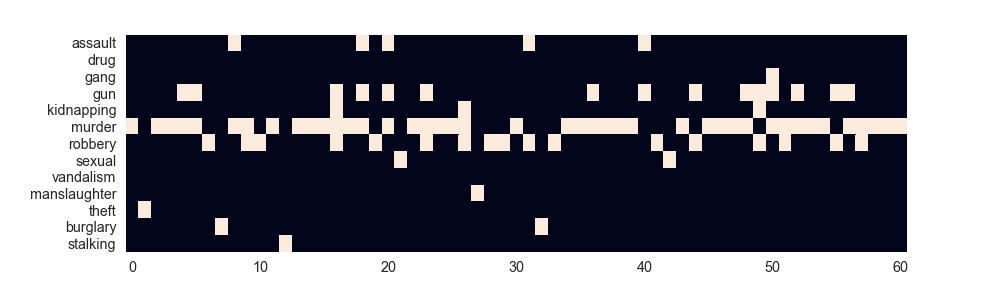}
        \caption{SSNMF}
        \label{fig:true_label}
  \end{subfigure}

  \begin{subfigure}[t]{0.8\textwidth}
        \includegraphics[scale=0.4]{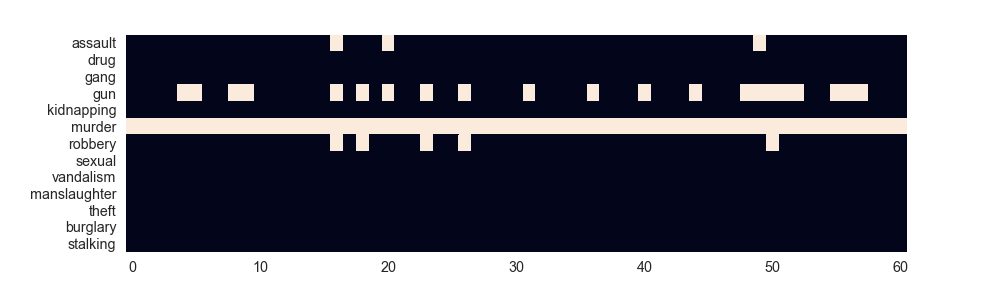}
        \caption{GSSNMF}
        \label{fig:label_reconstruction_SSNMF}
  \end{subfigure}

  \begin{subfigure}[t]{0.8\textwidth}
        \includegraphics[scale=0.4]{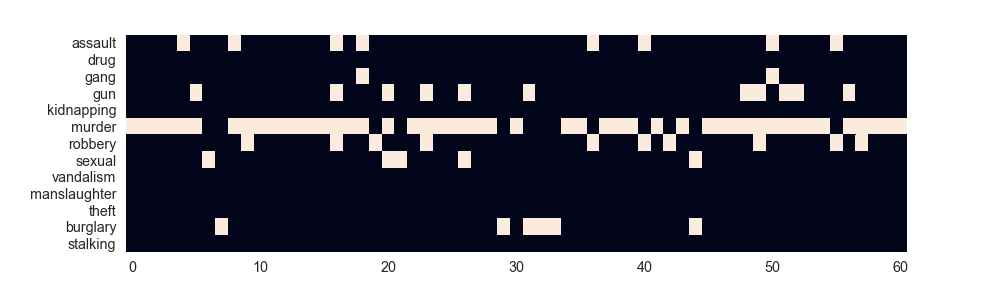}
        \caption{GSSNMF}
        \label{fig:label_reconstruction_GSSNMF}
  \end{subfigure}
  
  \caption{Actual and reconstructed crime testing label matrix using SSNMF and GSSNMF ($\mu$ = 0.0011, $\lambda$ = 0.0007), where a light pixel indicates that the case is assigned to the corresponding crime label on the y-axis, while a dark pixel indicates no assignment.}
    \label{fig:label_reconstruction}
\end{figure}

We can see that, in general, when incorporating the extra information from seed words, GSSNMF has a better Macro F1-score than SSNMF. As an example, we extract the reconstructed testing label matrix by SSNMF and GSSNMF along with the actual testing label matrix from a single trial. The matrices are visualized in Figure \ref{fig:label_reconstruction}. As we can see from the actual testing label matrix, \textit{murder} is a major label. Without the extra information from seed words, 
SSNMF tends to focus on the major label, leading to the trivial solution of classifying all cases as \textit{murder}; however, through user-specified seed words, GSSNMF can better evaluate the assignment of other labels, achieving an improved classification accuracy by the Macro F1-Score. 

\subsection{GSSNMF for Topic Modeling} \label{subsec:exp_tm}

In this section, we test the performance of GSSNMF for topic modeling by comparing it with Guided NMF on the CIP AOB dataset. Specifically, we conduct experiments for the range of rank identified in Section \ref{subsec:data}, running tests for various values of $\lambda$ and $\mu$ for each rank. To measure the effectiveness of the topics discovered by Guided NMF and GSSNMF, we calculate the \textit{topic coherence score} defined in \cite{mimno-etal-2011-optimizing} for each topic. The coherence score $\mathcal{C}_i$ for each topic $i$ with $N$ most probable keywords is given by

\begin{equation}
        \mathcal{C}_i = \sum_{b=2}^N\sum_{\ell = 1}^{b-1} \log \frac{P(w_b^{(i)}, w_\ell^{(i)})+1}{P(w_\ell^{(i)})}.
\label{eq:topic_coh}
\end{equation}
In the above equation (\ref{eq:topic_coh}), $P(w)$ denotes the \textit{document frequency} of keyword $w$, which is calculated by counting the number of documents that keyword $w$ appears in at least once. $P(w, w^\prime)$ denotes the \textit{co-document frequency} of keyword $w$ and $w^\prime$, which is obtained by counting the number of documents that contain both $w$ and $w^\prime$. In general, the topic coherence score seeks to measure how well the keywords that define a topic make sense as a whole to a human expert, providing a means for consistent interpretation of accuracy in topic generation. A large positive $\mathcal{C}$ coherence score indicates that the keywords from a topic are highly cohesive, as judged by a human expert.

Since we are judging the performance of methods that generate multiple topics, we calculate coherence scores $\mathcal{C}$ for each topic that is generated by Guided NMF or GSSNMF and then take the average. Thus, our final measure of performance for each method is the \textit{averaged coherence score }$\mathcal{C}_{\avg}$:
\begin{equation}
    \mathcal{C}_{\avg} = \frac{\sum_{i=1}^k \mathcal{C}_i}{k},
\label{eq:c_avg}
\end{equation}
where $k$ is the number of topics (or rank) we have specified. 

As suggested by Figure \ref{fig:TM_elbow_plot}, a proper rank falls between $6$ and $9$. Starting with the generation of $6$ topics from the AOB dataset, we first find a range of $\lambda$ in which Guided NMF generates the highest mean $\mathcal{C}_{\avg}$ over $10$ independent trials. In our computations, we use the top $30$ keywords of each topic to generate each coherence score; then for each trial, we obtain an individual $\mathcal{C}_{\avg}$, allowing us to average the 10 $\mathcal{C}_{\avg}$ from the 10 trials into the mean $\mathcal{C}_{\avg}$. Based on the proper range of $\lambda$, we then choose a range of $\mu$ for our GSSNMF to incorporate the label information into topic generation. Again, for each pair of $(\lambda,\mu)$, we run $10$ independent trials of GSSNMF and calculate $\mathcal{C}_{\avg}$ for each trial to generate a mean $\mathcal{C}_{\avg}$. With these ranges in mind, we work towards the following goal: for a given ``best" $\lambda$ of Guided NMF, we improve topic generation performance by implementing GSSNMF with a ``best" $\mu$ that balances how much weight GSSNMF should place on the new information of predetermined labels for each document. We then repeat the same process for ranks $7$, $8$, and $9$, and plot the mean $\mathcal{C}_{\avg}$ against each $\lambda$ in Figure \ref{fig:TM_lambda_plot}. The corresponding choice of $\mu$ can be found in the Appendix \ref{app:TM_details_on_mu}. We can see that most of the time, for a given $\lambda$, we are able to find such $\mu$ that GSSNMF can generate a higher mean $\mathcal{C}_{\avg}$ than Guided NMF in topic modeling across various ranks. Ultimately, we also see that a GSSNMF result always outperforms even the highest-performing Guided NMF result. 

In Table \ref{tab:TM_topic_output}, we provide an example of the outputs of topic modeling from Guided NMF using $\lambda=0.4$ and from GSSNMF using $\lambda = 0.3$ and $\mu = 0.006$ for a rank of 7. Note that we output only the top $10$ keywords under each identified topic group for ease of viewing, but our coherence scores are measured using the top $30$ keywords. Thus, while the top $10$ probable keywords of the generated topics may look similar across the two methods, the coherence scores calculated from the top $30$ probable keywords reveal that GSSNMF produces more coherent topics as a whole in comparison to Guided NMF. Specifically, GSSNMF demonstrates an ability to produce topics with similar levels of coherence (as seen from the small variance in individual coherence scores $\mathcal{C}$ of each topic), while Guided NMF produces topics that may vary in level of coherence (as seen from the large variance in individual coherence scores $\mathcal{C}$ for each topic). This further illustrates that GSSNMF is able to use the additional label information to execute topic modeling with better coherence. 


\begin{figure}[H]
\centering
  \begin{subfigure}[t]{0.8\textwidth}
        \includegraphics[scale=0.625]{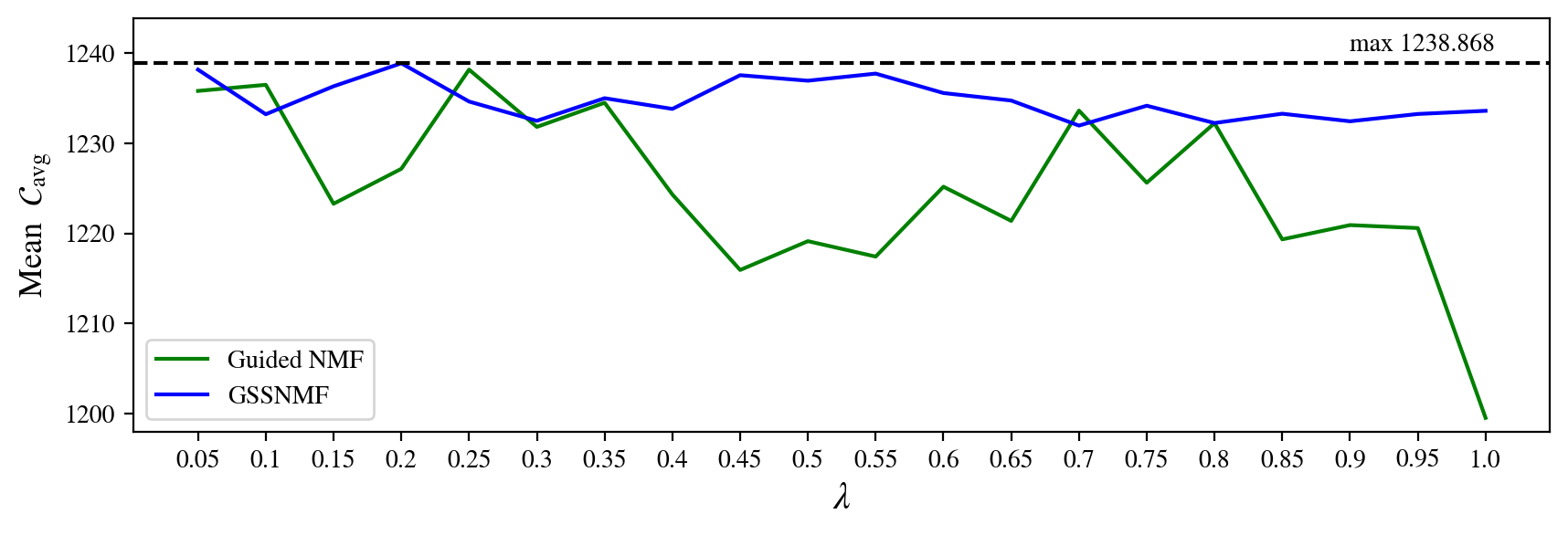}
        \caption{Rank $6$}
  \end{subfigure}

  \begin{subfigure}[t]{0.8\textwidth}
        \includegraphics[scale=0.625]{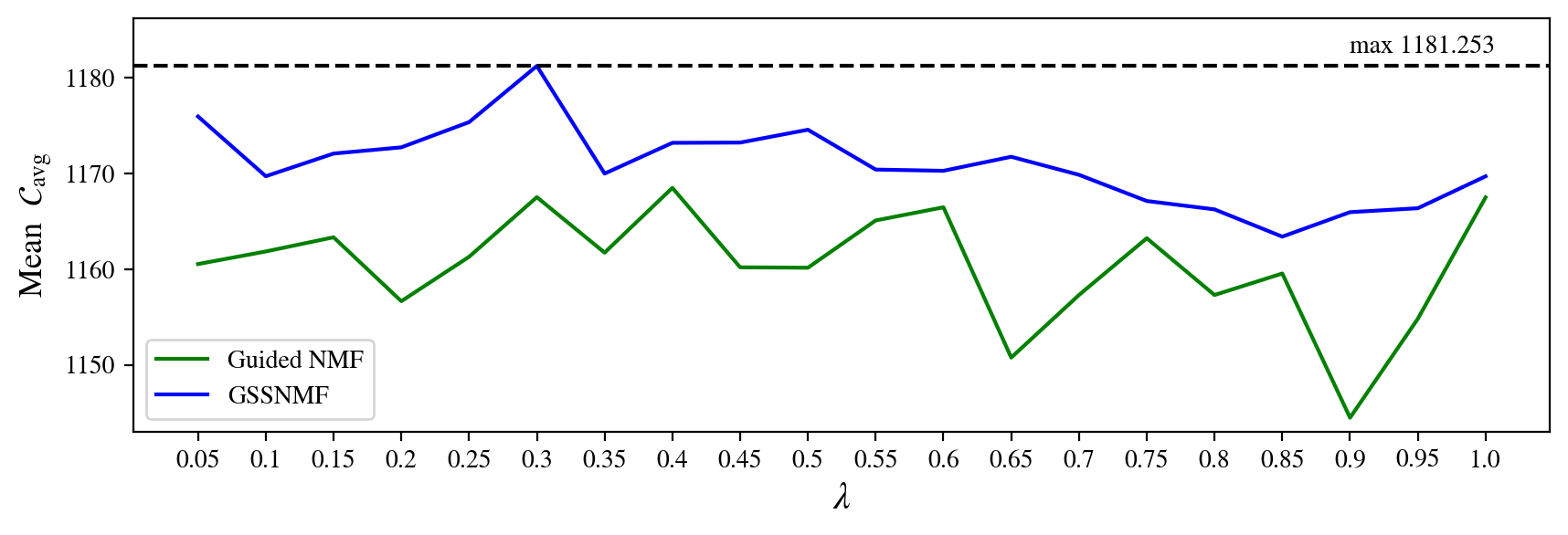}
        \caption{Rank $7$}
  \end{subfigure}
  
    \begin{subfigure}[t]{0.8\textwidth}
        \includegraphics[scale=0.625]{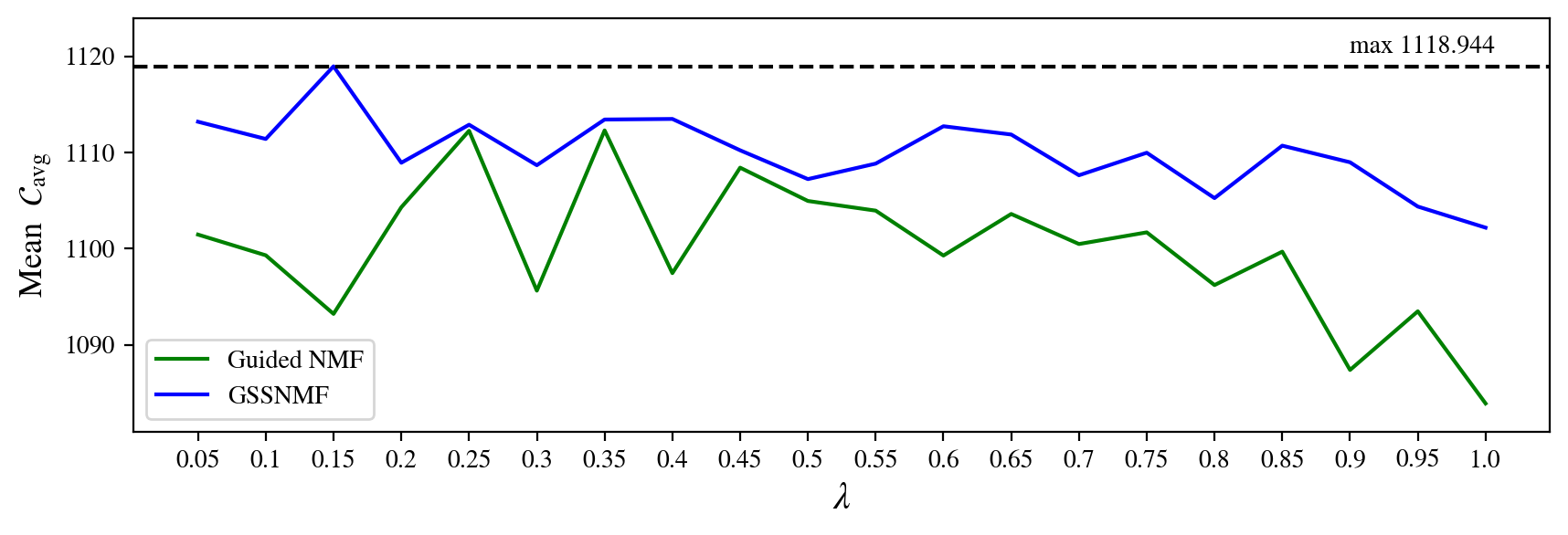}
        \caption{Rank $8$}
  \end{subfigure}
  
    \begin{subfigure}[t]{0.8\textwidth}
        \includegraphics[scale=0.625]{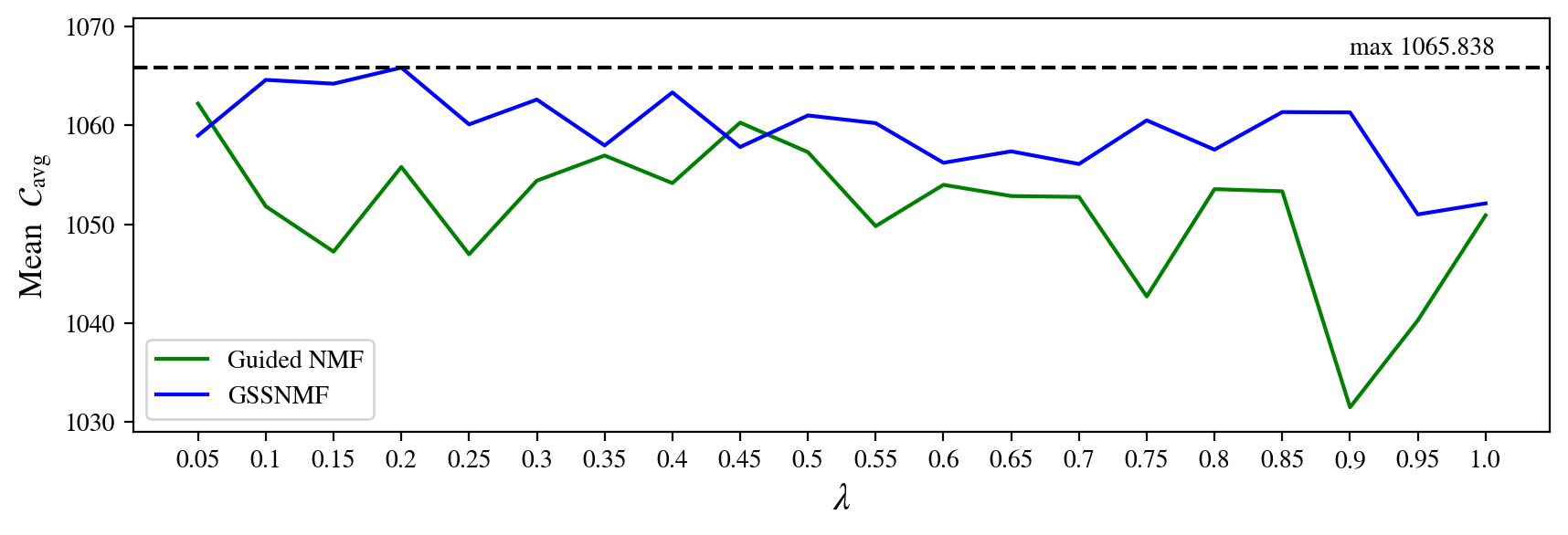}
        \caption{Rank $9$}
  \end{subfigure}
  \caption{Comparison of Guided NMF mean $\mathcal{C}_{\avg}$ score (over 10 independent trials) and highest GSSNMF mean $\mathcal{C}_{\avg}$ score (over 10 independent trials) for each $\lambda$ tested.} 
    \label{fig:TM_lambda_plot}
\end{figure}

\begin{table}[H]
    \centering
    \caption{Topic Modeling Results of Guided NMF and GSSNMF for Rank 7.}
    \begin{tabular}{c}
        \begin{tabular}{ccccccc}
            \multicolumn{7}{c}{Guided NMF Results ($\lambda = 0.4$)} \\
            \hline 
            Topic 1 & Topic 2 & Topic 3 & Topic 4 & Topic 5 & Topic 6 & Topic 7 \\
            \hline
            gang & burglari & murder & accomplic & gang & instruct & identif \\
            member & sexual & shot & corrobor & member & murder & eyewit \\
            crip & strike & hous & robberi & expert & manslaught & photo \\
            activ & admiss & detect & instruct & beer & lesser & lineup \\
            phone & instruct & vehicl & murder & estrada & theori & suggest \\
            murder & threat & phone & codefend & hispan & degre & suspect \\
            photo & object & apart & abet & tattoo & passion & photograph \\
            territori & discret & robberi & commiss & intent & abet & pack \\
            associ & impos & want & special & men & voluntari & procedur \\
            shot & sex & firearm & conspiraci & robberi & premedit & expert \\
            \hline
            \multicolumn{7}{c}{Coherence Score $\mathcal{C}$ per Topic:} \\
            \hline
            1112.94 & 1388.307 & 1290.817 & 921.023 & 1109.453 & 1123.185 & 1090.895 \\
            \hline
            \multicolumn{7}{c}{Averaged Coherence Score $\mathcal{C}_{\avg}$: 1148.089} \\
            \hline
        \end{tabular} \\
        \\
        \begin{tabular}{ccccccc}
            \multicolumn{7}{c}{GSSNMF Results ($\lambda = 0.3, \mu = 0.006$)} \\
            \hline 
            Topic 1 & Topic 2 & Topic 3 & Topic 4 & Topic 5 & Topic 6 & Topic 7 \\
            \hline
            murder & instruct & detect & identif & gang & gang & burglari \\
            accomplic & manslaught & phone & eyewit & member & member & strike \\
            corrobor & lesser & probat & photo & crip & expert & sexual \\
            vehicl & murder & waiver & lineup & associ & beer & robberi \\
            robberi & self & plea & suggest & expert & hispan & impos \\
            shot & passion & interrog & suspect & activ & shot & discret \\
            abet & theori & confess & photograph & intent & estrada & punish \\
            intent & voluntari & interview & pack & premedit & men & sex \\
            degre & heat & admiss & procedur & prove & tattoo & feloni \\
            hous & spont & transcript & reliabl & firearm & male & threat \\
            \hline
            \multicolumn{7}{c}{Coherence Score $\mathcal{C}$ per Topic:} \\
            \hline
            1215.826 & 1024.617 & 1188.975 & 1184.333 & 1180.321 & 1084.33 & 1281.146 \\
            \hline
            \multicolumn{7}{c}{Averaged Coherence Score $\mathcal{C}_{\avg}$: 1165.65} \\
            \hline
        \end{tabular} 
    \end{tabular}
    \label{tab:TM_topic_output}
\end{table}

\section{Conclusion and future works}

In this paper, we analyze the characteristics of SSNMF (Section \ref{subsec:subsec_ssnmf}) and Guided NMF (Section \ref{subsec:subsec_GuidedNMF}) concerning the tasks of classification and topic modeling. From these methods, we propose a novel NMF model, namely the GSSNMF, which combines characteristics of SSNMF and Guided NMF. SSNMF utilizes label information for classification; Guided NMF leverages user-specific seed words to guided topic content. Thus, to carry out classification and topic modeling simultaneously, GSSNMF uses additional label information to improve the coherence of topic modeling results while incorporating seed words for more accurate classification. Taking advantage of multiplicative updates, we provide a solver for GSSNMF and then evaluate its performance on real-life data. 

In general, GSSNMF is able to out-perform SSNMF on the task of classification. The extra information from the seed words contributes to a more accurate classification result. Specifically, SSNMF tends to focus on the most prevalent class label and classifies all documents into that class label. Unlike SSNMF, the additional information from choosing seed words as the class labels can help GSSNMF treat each class label equally and avoid the trivial solution of classifying every single document into the most prevalent class label. Additionally, GSSNMF is able to generate more coherent topics when compared to Guided NMF on the task of topic modeling. The extra information from the known label matrix can help GSSNMF better identify which documents belong to the same class. As a result, GSSNMF generates topics with higher and less variable coherence scores.

While there are other variants of SSNMF according to \cite{haddock2020semisupervised}, we developed GSSNMF only based on the standard Frobenius norm. In the future, we plan to make use of other comparable measures like the \textit{information divergence}, and derive a corresponding multiplicative updates solver. In addition, across all the experiments, we selected the parameters $\lambda$ and $\mu$, which put weight on the seed word matrix and label matrix respectively, based on the experimental results. In our continued work, we plan to conduct error analysis to determine how each parameter affects the other parameter and the overall approximation results. Particularly, for a given parameter $\lambda$ or $\mu$, we hope to identify an underlying, hidden relationship that allows us to quickly pick a matching $\mu$ or $\lambda$, respectively, that maximizes GSSNMF performance.

\section*{Acknowledgements}

The authors would like to thank Prof. Deanna Needell, Dr. Longxiu Huang, Joyce A. Chew, and Benjamin Jarman for their guidance on this project. The authors are also grateful for support received from the UCLA Computational and Applied Mathematics REU and NSF DMS \#2011140. 

\appendix
\section[\appendixname~\thesection]{GSSNMF Algorithm: Multiplicative Updates Proof} \label{app:GSSNMF_algorithm_proof}

\noindent
We begin with a corpus matrix $ \mX \in \mathbb{R}^{d\times n}_{\geq0}$, a seed matrix $\mY \in \mathbb{R}^{d\times s}_{\geq0}$, a label matrix $\mZ \in \mathbb{R}^{p\times n}_{\geq0}$, and a masking matrix $\mL \in \mathbb{R}^{p\times n}_{\geq0}$. From these, we hope to find dictionary matrix $\mW \in \mathbb{R}^{d\times k}_{\geq0}$, coding matrix $\mH \in \mathbb{R}^{k\times n}_{\geq0}$, and supervision matrices $\mB \in \mathbb{R}^{k\times s}_{\geq 0}$ and $\mC \in \mathbb{R}^{p\times k}_{\geq0}$ that minimize the loss function: 
\begin{align*}
    & F(\mW,\mH,\mB,\mC) \\
    =& \frac12 \|\mX-\mW\mH\|^2 + \frac{\lambda}{2} \|\mY - \mW\mB\|^2 +  \frac\mu 2 \|\mL\odot(\mZ-\mC\mH)\|^2 \\
    =& \frac12\tr[\mX\mX\tran - 2\mX\mH\tran \mW\tran + \mW\mH\mH\tran \mW\tran] + \frac \lambda 2 \tr[\mY\mY\tran -2\mY\mB\tran \mW\tran + \mW\mB\mB\tran \mW\tran] \\ 
    &~ + \frac\mu2 \tr[(\mL\odot \mZ) (\mL\tran \odot \mZ\tran) - 2 (\mL \odot \mZ)(\mL \tran \odot \mH \tran \mC \tran) + (\mL\odot \mC\mH)(\mL\tran\odot \mH\tran \mC\tran)].\\ 
\end{align*}

\noindent
Let $\alpha \in \mathbb{R}^{d\times k}, \beta \in \mathbb{R}^{k\times n}, \gamma \in \mathbb{R}^{k\times s}, \delta \in \mathbb{R}^{p\times k}$ be the Lagrange multipliers. The non-negative constraints on $\mW, \mH, \mB, \mC $ ensure that we must find solutions subject to
\begin{align*}
    \tr(\alpha \mW\tran), \ \tr(\beta \mH\tran), \ \tr(\gamma \mB\tran), \ \text{and} \ \tr(\delta \mC\tran),
\end{align*}
and with our objective to minimize, we arrive at the following Lagrangian function:
\begin{equation*}
\begin{aligned}
    \mathscr{L}(\mW,\mH,\mB,\mC,\alpha,\beta,\gamma,\delta) = F(\mW,\mH,\mB,\mC) + \tr(\alpha \mW\tran) + \tr(\beta \mH\tran) + \tr({\gamma} \mB\tran) + \tr(\delta \mC\tran).\\
\end{aligned}
\end{equation*}

\noindent
Taking the derivatives with respect to the matrices $\mW$, $\mB$, $\mH$ and $\mC$, we derive the First-Order conditions 
\begin{align*}
    \frac{\partial \mathscr L}{\partial \mW} &= -\mX\mH\tran + \mW\mH\mH\tran - \lambda \mY\mB\tran + \lambda \mW\mB\mB\tran + \alpha = 0, \\
    \frac{\partial \mathscr L}{\partial \mB} &= -\lambda \mW\tran \mY + \lambda \mW\tran \mW \mB + \gamma = 0, \\
    \frac{\partial \mathscr L}{\partial \mH} &= -\mW\tran \mX + \mW\tran \mW\mH - \mu \mC\tran(\mL\odot \mL\odot \mZ) + \mu \mC\tran(\mL\odot \mL\odot \mC\mH) + \beta = 0, ~\text{and} \\
    \frac{\partial \mathscr L}{\partial \mC} &= - \mu(\mL\odot \mL \odot \mZ)\mH\tran + \mu(\mL \odot \mL \odot \mC\mH)\mH\tran + \delta = 0. 
\end{align*}
With the Karush-Kuhn-Tucker conditions, we have complementary slackness condition 
\begin{align*}
    \alpha \odot \mW = \beta \odot \mH = \gamma \odot \mB = \delta \odot \mC = {0}.
\end{align*}
As such, we arrive at the following stationary equations:
\begin{gather*}
    \frac{\partial \mathscr L}{\partial \mW} \odot \mW = (-\mX\mH\tran + \mW \mH \mH\tran - \lambda \mY\mB\tran + \lambda \mW\mB\mC\mB\tran) \odot \mW + \underbrace{\alpha \odot \mW}_{=0} = 0, \\
    \frac{\partial \mathscr L}{\partial \mB} \odot \mB = (-\lambda \mW\tran \mY + \lambda \mW\tran \mW \mB) \odot \mB + \underbrace{\gamma \odot \mB}_{=0} = 0,\\
    \frac{\partial \mathscr L}{\partial \mH} \odot \mH = [-\mW\tran \mX + \mW\tran \mW \mH - \mu \mC\tran(\mL\odot \mL\odot \mZ) + \mu \mC\tran(\mL\odot \mL\odot \mC\mH)] \odot \mH + \underbrace{\beta \odot \mH}_{=0} = 0, \\
    \frac{\partial \mathscr L}{\partial \mC} \odot \mC = [- \mu(\mL\odot \mL \odot \mZ)\mH\tran + \mu(\mL \odot \mL \odot \mC\mH)\mH\tran]\odot \mC + \underbrace{\delta\odot \mC}_{=0} = 0.
\end{gather*}
With these, we derive the following updates: 
\begin{align*}
    \mW &\leftarrow \mW\odot \frac{\mX\mH\tran +\lambda \mY\mB\tran}{\mW\mH\mH\tran + \lambda \mW\mB\mB\tran}\;, \\
    \mB &\leftarrow \mB\odot \frac{\mW\tran \mY}{\mW\tran \mW\mB}\;, \\
    \mH &\leftarrow \mH\odot \frac{\mW\tran \mX + \mu \mC\tran (\mL\odot \mL \odot \mZ)}{\mW\tran \mW\mH + \mu \mC\tran(\mL\odot \mL\odot \mC\mH)}\;, \\
    \mC &\leftarrow \mC\odot \frac{(\mL\odot \mL \odot \mZ)\mH\tran}{(\mL \odot \mL \odot \mC\mH)\mH\tran}.
\end{align*}

\section[\appendixname~\thesection]{GSSNMF Topic Modeling: Details on $\mu$ Values} \label{app:TM_details_on_mu}

For each rank examined, we include tables in Table \ref{tab:appendix_mu_scores} of the mean averaged coherence scores (mean $\mathcal{C}_{\avg}$), defined by Equation \ref{eq:c_avg} in Section \ref{subsec:exp_tm}, 
corresponding to each $\lambda$ and its best-performing $\mu$ of the Guided NMF and GSSNMF methods.
These are the values that generate Figure \ref{fig:TM_lambda_plot} in Section \ref{subsec:exp_tm}.

\begin{table}[H]
    \centering
    \caption{Mean of averaged coherence scores from 10 independent trials (mean $\mathcal{C}_{\avg}$) of Guided NMF and GSSNMF given $\lambda$ and the best-performing $\mu$ for each $\lambda$, by rank.}
    \begin{tabular}{cc}
        \begin{tabular}{cccc}
            \hline
            \multicolumn{4}{c}{Rank 6} \\
            \hline
            \multicolumn{2}{c}{Guided NMF} & \multicolumn{2}{c}{GSSNMF} \\
            \hline
            $\lambda$ & Mean $\mathcal{C}_{\avg}$ & $\mu$ & Mean $\mathcal{C}_{\avg}$ \\
            \hline
            0.05 & 1235.799 & 0.017 & 1238.159 \\
            0.1 & 1236.479 & 0.006 & 1233.213 \\
            0.15 & 1223.286 & 0.011 & 1236.314 \\
            0.2 & 1227.161 & 0.01 & 1238.868 \\
            0.25 & 1238.161 & 0.006 & 1234.622 \\
            0.3 & 1231.811 & 0.014 & 1232.488 \\
            0.35 & 1234.488 & 0.019 & 1234.992 \\
            0.4 & 1224.315 & 0.015 & 1233.806 \\
            0.45 & 1215.942 & 0.002 & 1237.543 \\
            0.5 & 1219.138 & 0.009 & 1236.933 \\
            0.55 & 1217.428 & 0.019 & 1237.721 \\
            0.6 & 1225.177 & 0.009 & 1235.567 \\
            0.65 & 1221.387 & 0.011 & 1234.732 \\
            0.7 & 1233.618 & 0.014 & 1231.956 \\
            0.75 & 1225.619 & 0.019 & 1234.159 \\
            0.8 & 1232.224 & 0.014 & 1232.239 \\
            0.85 & 1219.339 & 0.02 & 1233.272 \\
            0.9 & 1220.92 & 0.002 & 1232.433 \\
            0.95 & 1220.591 & 0.0 & 1233.243 \\
            1.0 & 1199.539 & 0.01 & 1233.594 \\
        \end{tabular}
        &  
        \begin{tabular}{cccc}
            \hline
            \multicolumn{4}{c}{Rank 7} \\
            \hline
            \multicolumn{2}{c}{Guided NMF} & \multicolumn{2}{c}{GSSNMF} \\
            \hline
            $\lambda$ & Mean $\mathcal{C}_{\avg}$ & $\mu$ & Mean $\mathcal{C}_{\avg}$ \\
            \hline
            0.05 & 1160.539 & 0.004 & 1175.973 \\
            0.1 & 1161.856 & 0.008 & 1169.722 \\
            0.15 & 1163.338 & 0.004 & 1172.093 \\
            0.2 & 1156.654 & 0.006 & 1172.746 \\
            0.25 & 1161.297 & 0.013 & 1175.38 \\
            0.3 & 1167.535 & 0.006 & 1181.253 \\
            0.35 & 1161.732 & 0.018 & 1170.002 \\
            0.4 & 1168.508 & 0.004 & 1173.216 \\
            0.45 & 1160.197 & 0.01 & 1173.244 \\
            0.5 & 1160.156 & 0.002 & 1174.586 \\
            0.55 & 1165.1 & 0.017 & 1170.422 \\
            0.6 & 1166.476 & 0.019 & 1170.292 \\
            0.65 & 1150.758 & 0.017 & 1171.758 \\
            0.7 & 1157.302 & 0.008 & 1169.872 \\
            0.75 & 1163.242 & 0.008 & 1167.135 \\
            0.8 & 1157.301 & 0.017 & 1166.257 \\
            0.85 & 1159.55 & 0.005 & 1163.406 \\
            0.9 & 1144.468 & 0.017 & 1165.967 \\
            0.95 & 1154.814 & 0.018 & 1166.376 \\
            1.0 & 1167.511 & 0.011 & 1169.715 \\
        \end{tabular}\\
        \\
        \begin{tabular}{cccc}
            \hline
            \multicolumn{4}{c}{Rank 8} \\
            \hline
            \multicolumn{2}{c}{Guided NMF} & \multicolumn{2}{c}{GSSNMF} \\
            \hline
            $\lambda$ & Mean $\mathcal{C}_{\avg}$ & $\mu$ & Mean $\mathcal{C}_{\avg}$ \\
            \hline
            0.05 & 1101.472 & 0.013 & 1113.19 \\
            0.1 & 1099.322 & 0.007 & 1111.402 \\
            0.15 & 1093.238 & 0.014 & 1118.944 \\
            0.2 & 1104.311 & 0.017 & 1108.933 \\
            0.25 & 1112.251 & 0.018 & 1112.88 \\
            0.3 & 1095.652 & 0.01 & 1108.676 \\
            0.35 & 1112.29 & 0.003 & 1113.42 \\
            0.4 & 1097.472 & 0.018 & 1113.481 \\
            0.45 & 1108.421 & 0.001 & 1110.225 \\
            0.5 & 1104.963 & 0.015 & 1107.234 \\
            0.55 & 1103.96 & 0.02 & 1108.842 \\
            0.6 & 1099.296 & 0.012 & 1112.723 \\
            0.65 & 1103.613 & 0.004 & 1111.863 \\
            0.7 & 1100.494 & 0.018 & 1107.633 \\
            0.75 & 1101.705 & 0.003 & 1109.97 \\
            0.8 & 1096.231 & 0.006 & 1105.261 \\
            0.85 & 1099.701 & 0.017 & 1110.706 \\
            0.9 & 1087.417 & 0.001 & 1108.98 \\
            0.95 & 1093.507 & 0.013 & 1104.385 \\
            1.0 & 1083.938 & 0.019 & 1102.194 \\
        \end{tabular}
        &  
        \begin{tabular}{cccc}
            \hline
            \multicolumn{4}{c}{Rank 9} \\
            \hline
            \multicolumn{2}{c}{Guided NMF} & \multicolumn{2}{c}{GSSNMF} \\
            \hline
            $\lambda$ & Mean $\mathcal{C}_{\avg}$ & $\mu$ & Mean $\mathcal{C}_{\avg}$ \\
            \hline
            0.05 & 1235.799 & 0.017 & 1238.159 \\
            0.1 & 1236.479 & 0.006 & 1233.213 \\
            0.15 & 1223.286 & 0.011 & 1236.314 \\
            0.2 & 1227.161 & 0.01 & 1238.868 \\
            0.25 & 1238.161 & 0.006 & 1234.622 \\
            0.3 & 1231.811 & 0.014 & 1232.488 \\
            0.35 & 1234.488 & 0.019 & 1234.992 \\
            0.4 & 1224.315 & 0.015 & 1233.806 \\
            0.45 & 1215.942 & 0.002 & 1237.543 \\
            0.5 & 1219.138 & 0.009 & 1236.933 \\
            0.55 & 1217.428 & 0.019 & 1237.721 \\
            0.6 & 1225.177 & 0.009 & 1235.567 \\
            0.65 & 1221.387 & 0.011 & 1234.732 \\
            0.7 & 1233.618 & 0.014 & 1231.956 \\
            0.75 & 1225.619 & 0.019 & 1234.159 \\
            0.8 & 1232.224 & 0.014 & 1232.239 \\
            0.85 & 1219.339 & 0.02 & 1233.272 \\
            0.9 & 1220.92 & 0.002 & 1232.433 \\
            0.95 & 1220.591 & 0.0 & 1233.243 \\
            1.0 & 1199.539 & 0.01 & 1233.594 \\
        \end{tabular}\\
    \end{tabular}
    \label{tab:appendix_mu_scores}
\end{table}

\bibliography{bib1.bib}
\end{document}